\documentclass[runningheads]{llncs}
\usepackage{graphicx}
\usepackage{amsmath,amssymb} 
\usepackage{color}

\usepackage{color}

\usepackage[table]{xcolor} 

\usepackage{gensymb} 

\usepackage{tikz}
\usetikzlibrary{shapes}
\usetikzlibrary{arrows}
\usetikzlibrary{positioning}
\usetikzlibrary{fit}					
\usetikzlibrary{backgrounds}	
\usetikzlibrary{automata}
\usetikzlibrary{intersections}

\usepackage{pgfplots,pgfplotstable}
\pgfplotsset{compat=newest} 

\usetikzlibrary{pgfplots.groupplots}

\def\person#1#2{
    \begin{scope}[shift={#1},rotate around={{#2}:(0.5cm,0.5cm)}]
      \draw (0.2cm,0.5cm) [thick, fill=green!50,  densely dotted,opacity=0.6] circle (0.15cm);
      \draw (0.8cm,0.5cm) [thick, fill=green!50, densely dotted, opacity=0.6] circle (0.15cm);
      \draw (0.15cm,0.3cm) [thick, rounded corners=3.5, fill=green!50, densely dotted, opacity=0.8] rectangle (0.85cm,0.7cm);
      \draw (0.5cm,0.5cm) [thick, fill=black!100,  opacity=0.8] circle (0.18cm);
    \end{scope}
}

\def\guy#1#2{
    \begin{scope}[shift={#1},rotate around={{#2}:(0.5cm,0.5cm)}]
      \draw (0.2cm,0.5cm) [thick, fill=green!50] circle (0.15cm);
      \draw (0.8cm,0.5cm) [thick, fill=green!50] circle (0.15cm);
      \draw (0.15cm,0.3cm) [thick, rounded corners=3.5, fill=green!50] rectangle (0.85cm,0.7cm);
      \draw (0.5cm,0.5cm) [thick, fill=black!100] circle (0.18cm);
    \end{scope}
}

\tikzset{pics/person/.style 2 args={
   code={
      \pgfmathsetmacro{\personangle}{atan2(#2,#1)}
      \pgfmathsetlengthmacro{\personwidth}{0.4cm}
      \draw [rotate=\personangle] (0,\personwidth) [thick, fill=green!50] circle[radius=0.15cm];
      \draw [rotate=\personangle] (0,-\personwidth) [thick, fill=green!50] circle[radius=0.15cm];
      \draw [rotate=\personangle] (-0.2cm,-\personwidth) [thick, rounded corners=3.5, fill=green!50] rectangle (0.2cm,\personwidth);
      \draw [rotate=\personangle] (0.05cm,0) [thick, fill=black!10] circle (0.22cm);
      \draw [->, ultra thick] (0,0) -- (#1,#2); 
}}}

\tikzset{pics/guy/.style 2 args={
   code={
      \pgfmathsetmacro{\personangle}{atan2(#2,#1)}
      \pgfmathsetlengthmacro{\personwidth}{0.4cm}
      \draw [rotate=\personangle] (0,\personwidth) [thick, fill=green!50] circle[radius=0.15cm];
      \draw [rotate=\personangle] (0,-\personwidth) [thick, fill=green!50] circle[radius=0.15cm];
      \draw [rotate=\personangle] (-0.2cm,-\personwidth) [thick, rounded corners=3.5, fill=green!50] rectangle (0.2cm,\personwidth);
      \draw [rotate=\personangle] (0.05cm,0) [thick, fill=black!10] circle (0.22cm);
      \draw [->, ultra thick] (0,0) -- (#1,#2); 
}}}

\definecolor{mycyan}{RGB}{0,255,255}
\definecolor{manpurple}{RGB}{112,48,160}
\definecolor{manred}{RGB}{255,0,0}
\definecolor{manred2}{RGB}{238,130,238}
\definecolor{manblue}{RGB}{135,206,235}
\definecolor{manyellow}{RGB}{255,192,0}
\definecolor{colorgtgreen}{RGB}{0,128,0}

\tikzstyle{connector} = [->,thick]

\tikzset{
pics/person/.style 2 args={
   code={
      \pgfmathsetmacro{\personangle}{atan2(#2,#1)}
      \pgfmathsetlengthmacro{\personwidth}{0.4cm}
      \draw [rotate=\personangle] (0,\personwidth) [thick, fill=green!50] circle[radius=0.15cm];
      \draw [rotate=\personangle] (0,-\personwidth) [thick, fill=green!50] circle[radius=0.15cm];
      \draw [rotate=\personangle] (-0.2cm,-\personwidth) [thick, rounded corners=3.5, fill=green!50] rectangle (0.2cm,\personwidth);
      \draw [rotate=\personangle] (0.05cm,0) [thick, fill=black!10] circle (0.22cm);
      \draw [->, ultra thick] (0,0) -- (#1,#2); 
}}}

\begin{document}
\title{An RNN-based IMM Filter Surrogate} 

%
\titlerunning{An RNN-based IMM Filter Surrogate}

\author{Stefan Becker \orcidID{0000-0001-7367-2519} \and
Ronny Hug \orcidID{0000-0001-6104-710X} \and \\
Wolfgang H\"{u}bner \orcidID{0000-0001-5634-6324} \and Michael Arens \orcidID{0000-0002-7857-0332}}

%
\authorrunning{S. Becker, R. Hug, W. H\"{u}bner \and M. Arens}
%

\institute{Fraunhofer Institute for Optronics, \\System Technologies, and Image Exploitation (IOSB)\\
Gutleuthausstr. 1, 76275 Ettlingen, Germany\\
\email{stefan.becker@iosb.fraunhofer.de}\\}

\maketitle              
\begin{abstract}
The problem of varying dynamics of tracked objects, such as pedestrians, is traditionally tackled with approaches like the Interacting Multiple Model (IMM) filter using a Bayesian formulation. By following the current trend towards using deep neural networks, in this paper an RNN-based IMM filter surrogate is presented. Similar to an IMM filter solution, the presented RNN-based model assigns a probability value to a performed dynamic and, based on them, puts out a multi-modal distribution over future pedestrian trajectories. The evaluation is done on synthetic data, reflecting prototypical pedestrian maneuvers. 
\end{abstract}

\section{Introduction}
\label{sec:intro} 
The applications of pedestrian trajectory prediction cover a broad range from autonomous driving, robot navigation, smart video surveillance to object tracking. Traditionally, the task of object motion prediction is done by using a Bayesian formulation in approaches such as the Kalman filter \cite{kalman1960}, or nonparametric methods, such as particle filters \cite{Arulampalam_SP_2002}. Driven by the success of recurrent neural networks (RNNs) in modeling temporal dependencies in a variety of sequence processing tasks, such as speech recognition \cite{Graves_ICASSP_2013,chung2015recurrent} and caption generation \cite{Donahue_CVPR_2015,Xu_MLR_2015}, RNNs are increasingly utilized for object motion prediction \cite{alahi2016social,alahi2017learning,Hug_RFMI_2017,Hug_ITSC_2018,Becker_ECCVW_2018}. When relying on traditional approaches, the challenge of varying dynamics over time is commonly addressed with the Interacting Multiple Model (IMM) filter \cite{Blom_AC_1988}. The IMM filter is a well established approach to elegantly combine a set of candidate models into a single context by weighting each individual model. Each model corresponds to a specific motion pattern and contributes to the final state estimation depending on its current weight. According to the IMM filter solution, in this paper an RNN-based IMM filter surrogate is presented. On the one hand, the presented RNN-based model is able to also provide a confidence value for the performed dynamic and on the other hand can overcome some limitations of the classic IMM filter. The suggested RNN-encoder-decoder model generates the probability distribution over future pedestrian paths conditioned on a dynamic class. The model is based on the work of Deo and Trivedi \cite{Deo_IV_2018}. For the case study of freeway traffic, they used an two branch RNN-encoder-decoder network for vehicle maneuver and trajectory prediction. Since for vehicle applications an on-board lane estimation algorithm is mostly available, a stationary frame of reference, with the origin fixed at the vehicle being predicted, is used in their work. Although this makes the model independent of road curvature and independent of how vehicle tracks are obtained, it can not be applied without adjustments for pedestrian motion prediction. Thus, our RNN-based model infers like classical filters the current position and uses only a single RNN branch for encoding the maneuver class, the filtered position and the trajectory information. In the context of vehicle motion prediction, maneuver or rather dynamic classes can be better defined than for pedestrians. For example by changing or keeping the lane. Due to the dynamic behavior of pedestrians, the maneuver classes are here defined based on the deviation from a straight walking pedestrian. The presented network also extends the maneuver network of Deo and Trivedi \cite{Deo_IV_2018} with insights from the work of Becker et al. \cite{Becker_ECCVW_2018} to better adapt to pedestrian motions.\\ 
Moreover, this paper aims to highlight some relations between traditional multiple model approaches such as the IMM filter and the suggested RNN-based IMM filter surrogate. By combining the different views on maneuver predictions, this work contributes to an exploration of the connections between both problem formulations. The decoder uses the de-noised position estimate and a context vector, encoding the dynamic classes, to predict future positions. The analysis is done on synthetic data reflecting prototypical scenarios capturing pedestrians maneuvers.\\
In the following, a brief formalization of the problem and a description of the RNN-based model are provided. The achieved results are presented in section \ref{sec:eval}. Finally, a conclusion is given in section \ref{sec:conclusion}. 

\section{RNN-based IMM Filter Surrogate}
\label{sec:model}
The goal is to devise a model that can successfully predict future paths of pedestrians and represent alternating pedestrian dynamics, e.g. dynamics that can transition from a straight walking to a turning maneuver or stopping. Here, trajectory prediction is formally stated as the problem of predicting the future trajectories of a pedestrian, conditioned on its track history. Given an input sequence $\mathcal{Z} = \{ (x^{t},y^{t}) \in \mathbb{R}^2 | t = 1,\ldots,t_{obs} \}$ of $T_{obs}$ consecutive observed pedestrian positions $\vec{z}^t=(x^t,y^t)$ at time $t$ along a trajectory, the task is to generate a multi-modal prediction for the next $T_{pred}$ positions $\{ \vec{x}^{t+1}, \vec{x}^{t+2},\ldots, \vec{x}^{t+T_{pred}} \} $ and to filter the current position $\vec{x}^t=(x^t,y^t)$. One insight from the work Becker et al. \cite{Becker_ECCVW_2018} is that motion continuity is easier to express in offsets or velocities, because it takes considerably more modeling effort to represent all possible conditioning positions. In order to exploit scene-specific knowledge for trajectory prediction, additional use of the position information is required. When sufficient training samples from a particular scene are available, Hug et al.\cite{Hug_RFMI_2017} showed that RNN-based trajectory prediction models are able to capture spatially dependent behavior changes only from motion data. However, here the offsets are additionally used for conditioning the network $\mathcal{Z} = \{(x^{t},y^{t}, \delta^{t}_{x},\delta^{t}_{y}) \in \mathbb{R}^4 | t= 2,\ldots,t_{obs} \}$. Apart from the smaller modeling effort to represent conditioned offsets, the shift to offsets helps to prevent undefined states due to a limited data range \cite{Becker_ECCVW_2018} and it is easier to make better generalizations across datasets. Since, we analyze the model capabilities on synthetic data reflecting prototypical pedestrian maneuvers for a fixed scenario, the amount of training samples is not restricted. Thus, in order to localize in the reference system position information is used to estimate the true position. The future trajectory is denoted with $\mathcal{Y} = \{ (x^{t},y^{t}) \in \mathbb{R}^2 | t = t_{obs}+1,\ldots,t_{pred} \}$. The model estimates the conditional distribution $P(\mathcal{Y},\vec{x}^{t}|\mathcal{Z})$. 
In order to identify specific dynamics under $M$ desired maneuver classes (e.g. turning maneuvers, stopping and straight walking), this term can be given by:

\begin{equation}
  P(\mathcal{Y},\vec{x}^t|\mathcal{Z}) = \sum^{M}_{i=1} P_{\Theta}(\mathcal{Y},\vec{x}^t|m_{i},\mathcal{Z})P(m_{i}|\mathcal{Z})
\label{eq:equation_01} 
\end{equation}

Here, $\Theta=\{\Theta^{t_{obs}+1},\ldots, \Theta^{t_{pred}} \}$ are the parameters of a $L$ component Gaussian mixture model $\Theta^{t} = (\vec{\mu}^{t}_{l}, \Sigma^{t}_{l}, w^{t}_{l})_{l=1,\ldots, L}$. By adding the maneuver context in form of the posterior mode probability, $P(m_{i}|\mathcal{Z}) \overset{\wedge}{=} \alpha_{i}$ the analogy to the classic IMM filter becomes apparent. For an IMM filter, the mode probability is used to calculate the mixing probabilities to combine the set of chosen candidate models into a merged estimate. The time behavior of the basic filter set is modeled as a homogeneous (time invariant) Markov chain with a fixed transition probability matrix (TPM) $m_{ij} \overset{\wedge}{=} P(m^{t}_{i}|m^{t-1}_{j})$. Under the assumption that $M$ models describe the variation of the dynamics, the posterior density of the IMM filter can be written as follows: 

\begin{equation}
  P(\vec{x}^t|\mathcal{Z}) = \sum^{M}_{i=1} P_{\Theta_{IMM}}(\vec{x}^t|m_{i},\mathcal{Z})P(m_{i}|\mathcal{Z})
\label{eq:equation_02} 
\end{equation}

Here, $P_{\Theta_{IMM}}(\vec{x}^t|m_{i},\mathcal{Z})$ is in the context of an IMM filter a Gaussian distribution and $P(m_{i}|\mathcal{Z}) \overset{\wedge}{=} \alpha_{i}$ is the posterior mode probability for the IMM filter. As mentioned above, the transition between different dynamics is modeled as a first order Markov chain for an IMM filter. The law of total probability allows to compute new mode probabilities based on the transition probabilities.  
Given the current mode probabilities and transition probabilities, the mixing probabilities $\alpha_{i|j}$ for the mixing step of the IMM filter can be calculated. For each model $M_{i}$ and $M_{j}$, they are calculated as $\alpha^{t-1}_{i|j} = \frac{1}{\bar{c}_{j}} m_{ij} \alpha^{t-1}_{i}$ with a normalization factor $\bar{c}_{j} = \sum^{M}_{i=1}  m_{ij} \alpha^{t-1}_{i}$. Then, in the prediction stage, each filter is applied independently using the calculated mixed initial condition. Subsequently, the model probabilities are adapted according to the likelihood of each filter.\\

\textbf{RNN-IMM:} Whereas an explicit modeling of the switching behavior and the object dynamics of the IMM filter stands in contrast to an implicit dynamic encoding of an RNN-based approach. In order to provide an IMM filter surrogate, the proposed model also estimates mode probabilities and filters or rather de-noises the current position based on noisy observations $\mathcal{Z}$. By writing the conditional distribution $P(\mathcal{Y},\vec{x}|\mathcal{Z})$ of the RNN-based approach in form of equation \ref{eq:equation_01}, the desired estimates can be inferred from the hidden states of the RNN $\vec{h}$. This formulations does not require to set the parameters of the TPM matrix manually, which  is commonly done based on the mean sojourn time (the mean time an object stays in a motion type \cite{Schneider_GCPR_2013,BarShalom_book_2002}) or as stated in the work of Bar-Shalom \cite{BarShalom_book_2002}, an ad-hoc approach to fill the diagonals with values close to one. For the proposed RNN-based IMM filter surrogate (RNN-IMM), the basic architecture is a recurrent encoder-decoder model. The encoder takes the frame by frame input sequence $\mathcal{Z}$. The hidden state vector of the encoder is updated at each time step based on the previous hidden state and the current observation. The generated internal representation is used to predict mode probabilities ${\vec{\alpha}}^t$ at the current time step and ${\vec{x}}^t$. 
With embedding of the current observations, the encoder can be defined as follows:

 \begin{gather*}
 \vec{e}^{t}_{encoder} = \text{EMB}(\vec{z}^{t}; W_{ee} )  \\
 \vec{h}^{t}_{encoder} = \text{RNN}(\vec{h}^{t-1}_{encoder},\vec{e}^{t}_{encoder}; W_{encoder} )  \\ 
 \hat{\vec{x}}^t,\vec{\alpha}^t_{logits} = \text{MLP}(\vec{h}^{t}_{encoder}; W_{en})\\
\hat{\vec{\alpha}}^t = \frac{\exp{(\vec{\alpha}^t_{logits}})}{\sum^{M}_{j=1} \exp{(\alpha^t_{logits, j}}) }\\ 
\label{eq:encoder} 
\end{gather*}

Here, $\text{RNN}(\cdot)$ is the recurrent network, $\vec{h}$ the hidden state of the RNN, $\text{MLP}(\cdot)$ the multilayer perceptron, and $\text{EMB}(\cdot)$ an embedding layer. $W$ represents the weights and biases of the MLP, EMB or respectively RNN. The final state of the encoder can be expected to encode information about the track histories. For generating a trajectory distribution over dynamic modes, the encoder hidden state is appended to a one-hot encoded vector corresponding to specific maneuvers and the filtered current position. Instead of only filtering the position, the encoder could also be used to parametrize a mixture density output layer (MDL). The decoder of the model can be defined as follows:

\begin{gather*}
\vec{h}^{t}_{decoder}  = \text{RNN}(\vec{h}^{t-1}_{decoder}[\vec{h}^{t}_{encoder}], \hat{\vec{x}}^t ,\vec{\alpha}^t ; W_{decoder} )\\
\hat{\mathcal{Y}} = \{ (\hat{\vec{\mu}}^{t}_{l}+  \hat{\vec{x}}^{t_{obs}}, \hat{\Sigma}^{t}_{l}, \hat{w}^{t}_{l})  | t= t_{obs}+1,\ldots,t_{pred} \} = \text{MLP}(\vec{h}^{t}_{decoder}; W_{de})
\label{eq:dencoder} 
\end{gather*}

The decoder is used to parametrize a mixture density output layer (MDL) or rather $\Theta$ directly for several positions in the future (one distribution for every time step). Nevertheless, the overall RNN-IMM uses the trajectory prediction and dynamic classification jointly, the loss function for training is split into three parts.

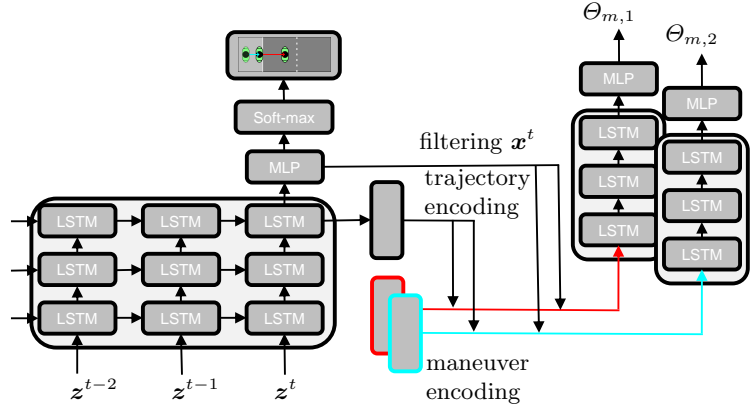
\begin{figure}[h!]
  \begin{center}
	\begin{tikzpicture}	
	\node (image1) at (0,0.) {\includegraphics[width=\textwidth]{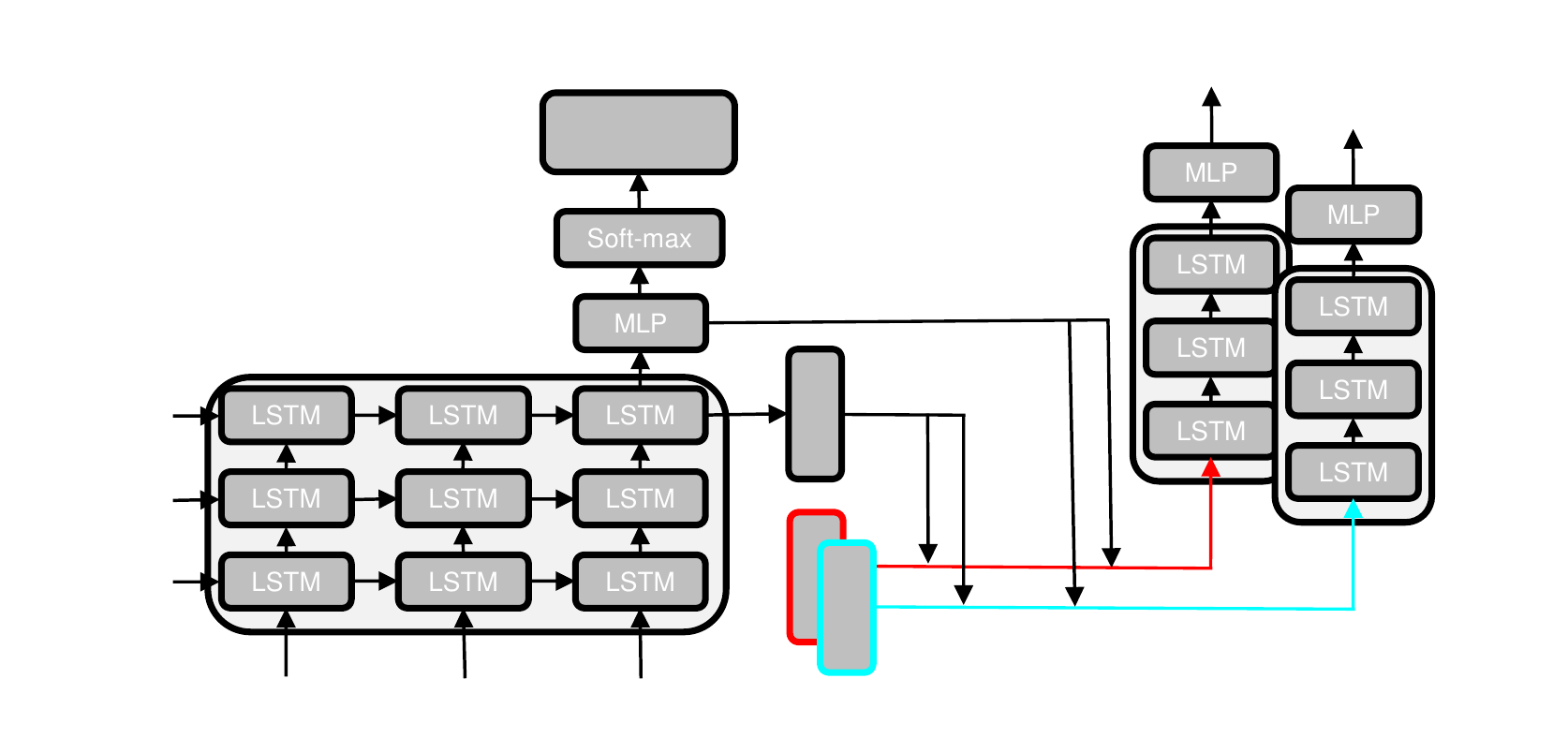}};
	\node (image2) at (-1.12,1.85) { \includegraphics[width=0.1\textwidth]{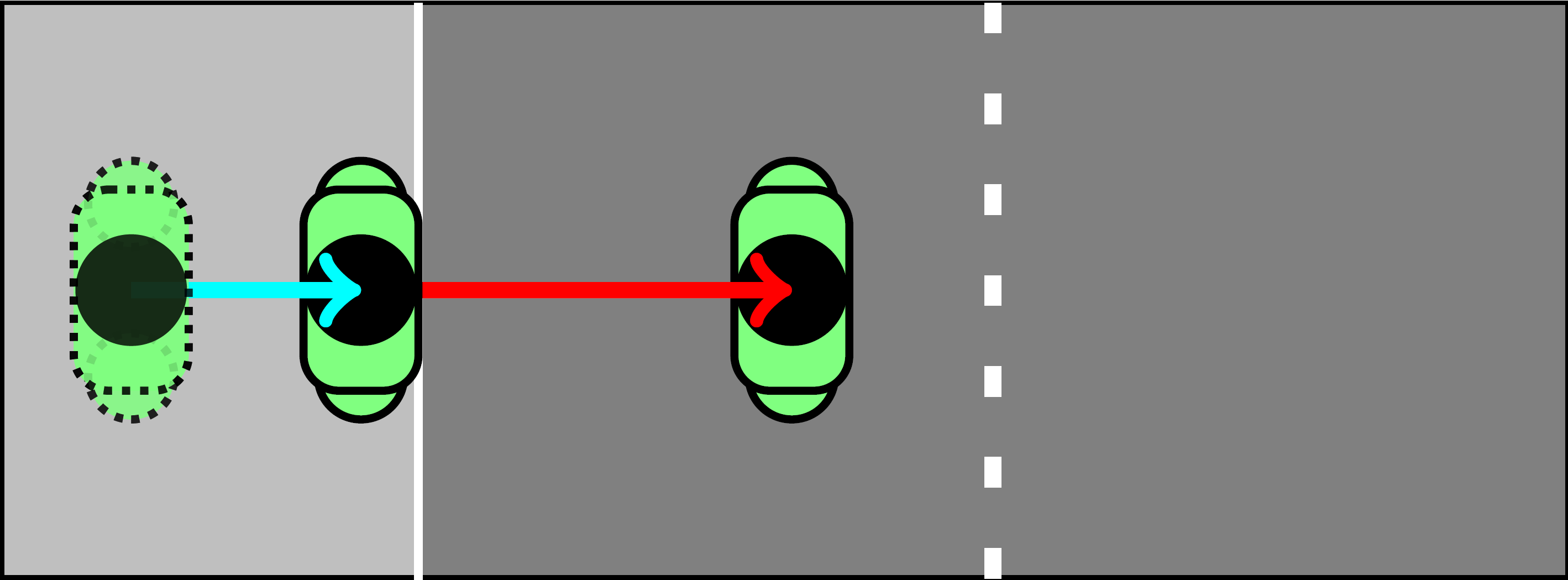}};
	\node[align=center] (note0) at (1.45,0.7){\small filtering $\vec{x}^{t}$};
	\node[align=left] (note1) at (1.45,.){\small trajectory \\ encoding};
	\node[align=left] (note2) at (1.45,-2.5){\small maneuver \\ encoding};
	\node[align=left] (note0) at (-1.1,-2.6){\small $\vec{z}^{t}$};
	\node[align=left] (note0) at (-2.3,-2.6){\small $\vec{z}^{t-1}$};
	\node[align=left] (note0) at (-3.65,-2.6){\small $\vec{z}^{t-2}$};
	\node[align=left] (note0) at (3.2,2.4){\small $\Theta_{m,1}$};
	\node[align=left] (note0) at (4.3,2.1){\small $\Theta_{m,2}$};
	\end{tikzpicture}
	\end{center}
	\caption{\label{fig:red} Visualization of the RNN-based IMM filter surrogate (RNN-encoder-decoder network) for jointly predicting specific dynamic probabilities and corresponding future distributions of trajectory positions. The encoder predicts the dynamic probabilities and the filtered position for the current time step. The decoder uses the context vector and the position estimate to predict future pedestrian locations.} 
\end{figure}

Dynamic classification is trained to mimimize the sum of cross-entropy losses of the different $M$ motion model classes:

\begin{equation}
  \mathcal{L}(\mathcal{Z} )_{maneuver} = -\sum^{M}_{j=1}  \alpha^t_{j,GT}  \log ( \hat{\alpha}^t_{j} )
\label{eq:loss_funtion_maneuver} 
\end{equation} 

Additionally, the encoder is trained by minimizing the filtering loss $\mathcal{L}(\mathcal{Z})_{filter}$ in form of the mean squared error to the ground truth current pedestrian locations. In case the encoder should generate the parameter of a mixture of Gaussian or single Gaussian distribution, the negative log likelihood for the ground truth pedestrian locations can be minimized. Finally, the complete encoder-decoder is trained by minimizing the negative log likelihood for the ground truth future pedestrian locations conditioned under the performed maneuver class. The context vector is appended with the ground truth values of the dynamic model or maneuver classes for each training trajectory. This results in the following loss function:

\begin{align}
\begin{split}
\mathcal{L}(\mathcal{Z})_{pred} &=  - \log ( P_{\Theta}( \hat{\mathcal{Y}}|m_{GT},\mathcal{Z})P(m_{GT}|\mathcal{Z}))\\
	\mathcal{L}(\mathcal{Z})_{pred} &= \sum_{t=t_{obs}+1}^{t_{pred}} -\log ( \sum^{L}_{l=1} \hat{w}_{l}^t \mathcal{N}( \vec{x}^{t} | \hat{\vec{\mu}}^{t}_{l}+ \vec{x}^{t_{obs}}, \hat{\Sigma}^{t}_{l}; m_{GT}) ) \\
\end{split}
\end{align} 

The overall architecture is visualized in figure \ref{fig:red}. The context vector combines the encoding of the track history with the encoding of the alternating dynamic classes. Together with the filtered position, it is used as input for the decoder. 

\section{Data Generation and Evaluation}
\label{sec:eval}

\vspace{-1.cm}
\begin{figure}[!ht]
\begin{center}
\vspace{-1.cm}
\begin{tabular}{cc}
   \begin{tikzpicture}    
		\draw[fill=gray] (0,0) -- (0,2) -- (4,2) -- (4,0)  -- cycle;
		\draw[fill=gray!50] (4,0) -- (4,2) -- (5.45,2) -- (5.45,0)  -- cycle;
		\draw[ultra thick,loosely dashed,white] (2,0) -- (2,4);
		\draw[thick,white] (4,0) -- (4,3); 				
		\guy{(2.2, 0.5)}{90};		 
		\draw (5.,1.0) [->, ultra thick, manred] to [out=180,in=0] (2.7,1.0);
		\person{(4.5,0.5)}{90};	
  \end{tikzpicture}  &
	\begin{tikzpicture}  
		\draw[fill=gray] (0,0) -- (0,2) -- (4,2) -- (4,0)  -- cycle;
		\draw[fill=gray!50] (4,0) -- (4,2) -- (5.45,2) -- (5.45,0)  -- cycle;
		\draw[ultra thick,loosely dashed,white] (2,0) -- (2,4);
		\draw[thick,white] (4,0) -- (4,3);
		\guy{(3.7, 0.5)}{90};		
		\draw (5.,1.0) [->, ultra thick,mycyan] to [out=180,in=0] (4.2,1.0);		
		 \person{(4.5,0.5)}{90};
  \end{tikzpicture}	
	\\
	\begin{tikzpicture}
		\node (image1) at (0,0.) { \includegraphics[width=0.45\textwidth]{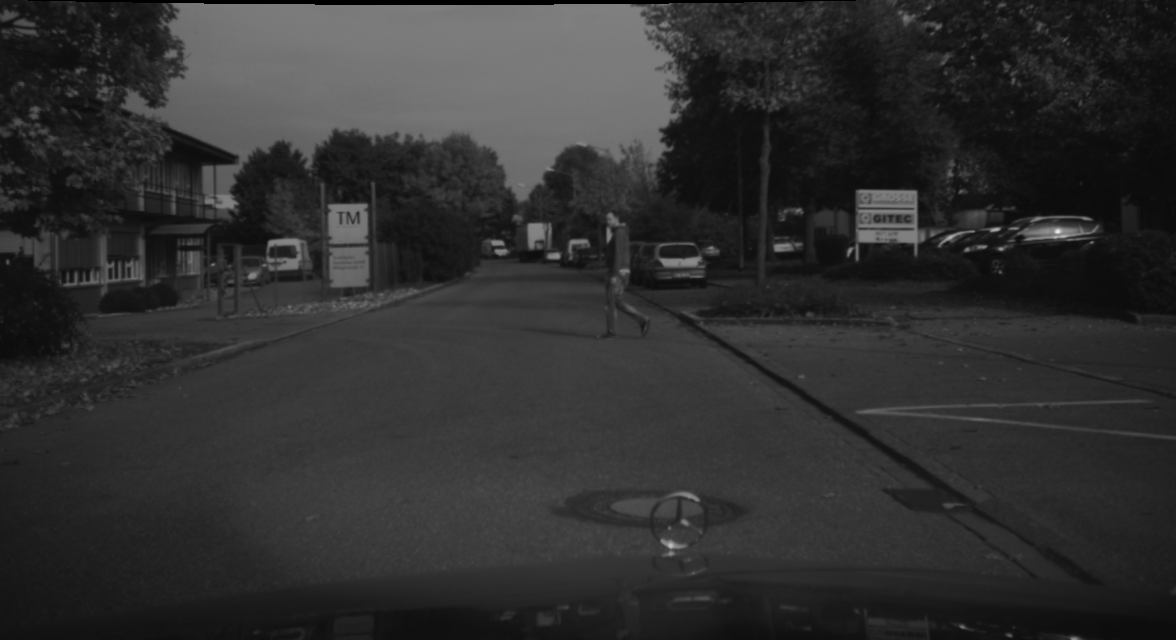}};
		\draw [->, ultra thick, manred] (2.2,-0.1) to [out=180, in=0] (.15,0.2);
		\draw[manred,rounded corners] (.0,0.52) rectangle (.3,-0.1);
	\end{tikzpicture}& 
	\begin{tikzpicture}
		\node (image2) at (0,0.0) {\includegraphics[width=0.45\textwidth]{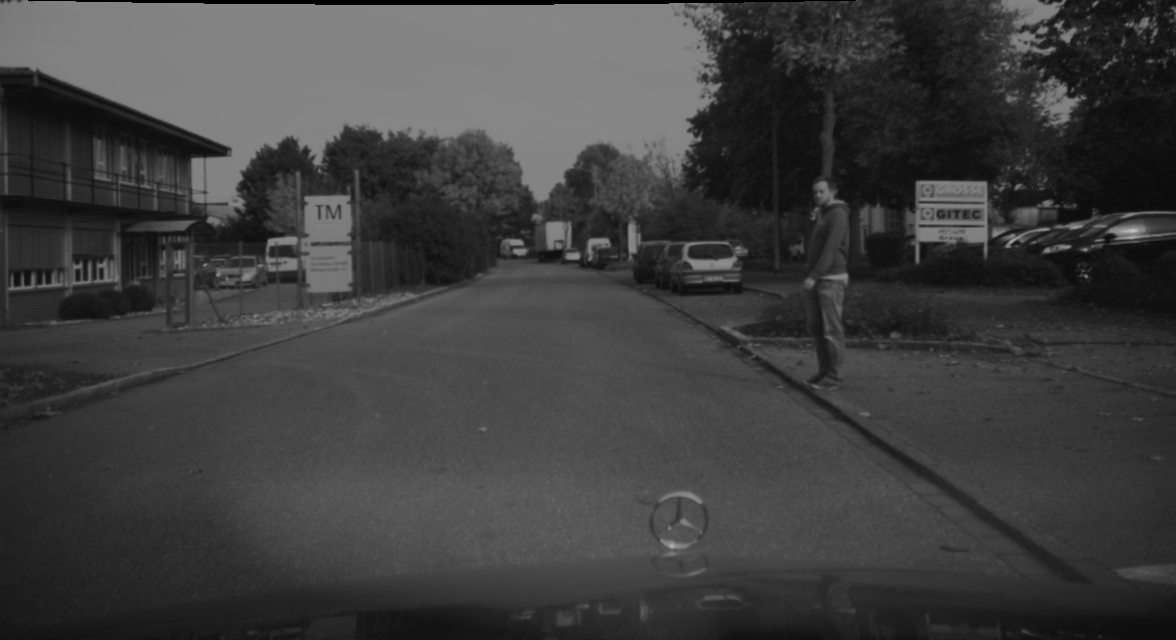}};			
		\draw [->, ultra thick, mycyan] (2.2,0.) to [out=180, in=0] (1.1,0.2);
		\draw[mycyan,rounded corners] (.9,0.7) rectangle (1.3,-0.3);	
	\end{tikzpicture}
	\\
		\begin{tikzpicture}
		\node (image1) at (0,0.) { \includegraphics[height=0.23\textwidth,width=0.45\textwidth]{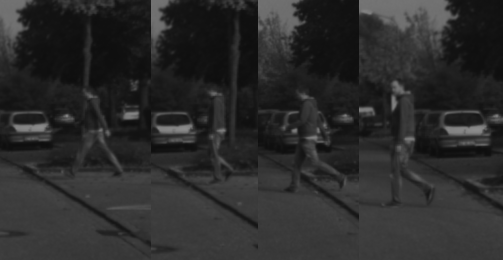}};
		\draw[manred,ultra thick,rounded corners] (1.4,0.6) rectangle (2.,-0.8);
		\draw[manred,ultra thick,rounded corners] (.35,0.5) rectangle (1.,-0.7);
		\draw[manred,ultra thick,rounded corners] (-.55,0.5) rectangle (-0.15,-0.6);
		\draw[manred,ultra thick,rounded corners] (-2.,0.5) rectangle (-1.4,-0.55);	
	\end{tikzpicture}& 
	\begin{tikzpicture}
		\node (image2) at (0,0.0) {\includegraphics[height=0.23\textwidth,width=0.45\textwidth]{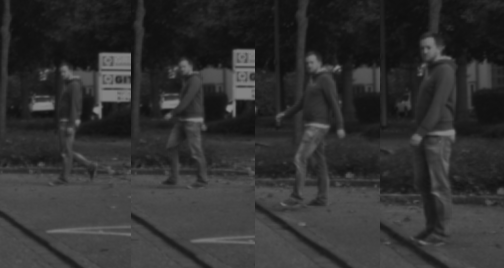}};
		\draw[mycyan,ultra thick,rounded corners] (1.7,1.1) rectangle (2.25,-1.2);
		\draw[mycyan,ultra thick,rounded corners] (.25,0.9) rectangle (1.,-0.8);
		\draw[mycyan,ultra thick,rounded corners] (-1.0,0.8) rectangle (-0.45,-0.6);
		\draw[mycyan,ultra thick,rounded corners] (-2.2,0.8) rectangle (-1.65,-0.55);			
	 	\end{tikzpicture}
\end{tabular}
\end{center}
\caption{\label{fig:crossing_stopping} Illustration of typical pedestrian motions. The above images depict the two chosen maneuver classes of straight walking or rather crossing and stopping. The images on the left show a person crossing the street. The images on the right show a person changing from walking to standing at the curbside of the street. In particular changing from straight walking to stopping \cite{Schneider_GCPR_2013}.}   
\end{figure}
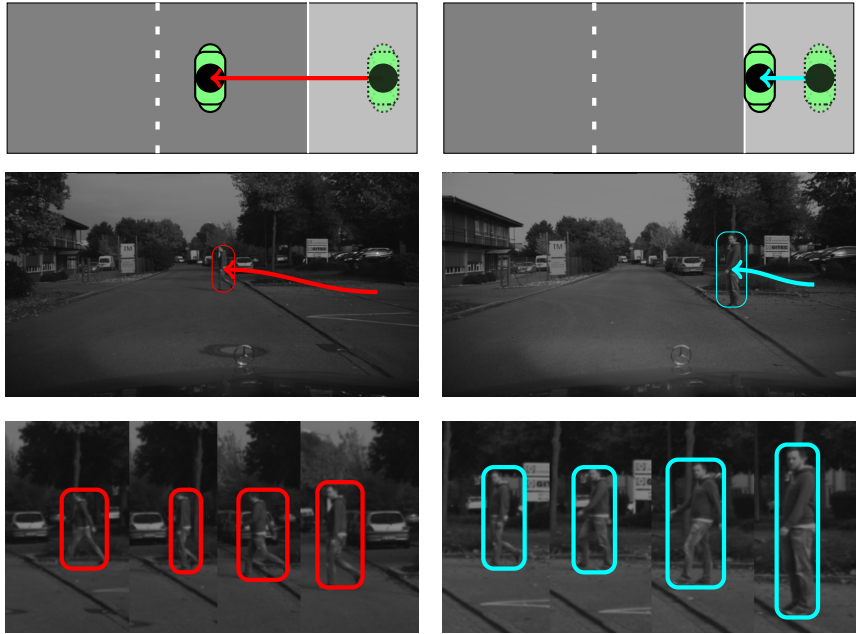

This section consists of a brief evaluation of the proposed RNN-IMM. The evaluation is concerned with verifying the overall viability of the approach in maneuver situations. For initial results, a synthetic test condition is used in order to gain insight into the model behavior in different typical pedestrian motion types. A prototypical maneuver performed by a pedestrian, which has important implications for the field of intelligent vehicles and video surveillance is a stopping or deceleration maneuver.\\

\textbf{Data Generation and Reference Methods:} For the first mentioned context of intelligent vehicles, Schneider et al. \cite{Schneider_GCPR_2013} performed a comparative study on recursive Bayesian filters for pedestrian path prediction at short time horizons (below $2$ seconds). They applied different filters on typical pedestrian motion types. Although, the comparison was done on the Daimler path prediction dataset, we evaluate on synthetic data but make use of the provided real data to capture a similar condition.
Firstly, the Daimler path prediction dataset provides only a maximum amount of $23$ sequences for single motion types. As mentioned before, in order to avoid problems such as a limited number of training samples and to gain some insights into a controlled setup, synthetic data is used. Secondly, the location information is biased in the dataset. Since recursive Bayesian filters make in their standard formulation no use of the spatial context of a scene, this does not harm their mutual comparison. However, RNN-based prediction networks are able to capture spatially dependent behavior changes \cite{Hug_RFMI_2017}, thus a fair comparison is difficult to achieve. The evaluation on the Daimler dataset is done in an ego-motion compensated reference system. The frame rate of the camera system inside the recording vehicle is $16$ fps and it is taken over accordingly for our experiments. The pedestrians change their behavior abruptly. Therefore, the sensible time horizons are short. Here, $8$ ($0.5$ seconds) consecutive positions are observed, before predicting the next $8$ ($0.5s$ seconds), $12$ ($0.75$ seconds) and $16$ ($1$ second).\\
For generating synthetic trajectories of a basic maneuvering pedestrian, random agents are sampled from a Gaussian distribution according to a preferred pedestrian walking speed \cite{Teknomo_2016} ($\mathcal{N}(1,38 m,0.37m)$) from the distribution of starting positions of the corresponding Daimler dataset sequences. During a single trajectory simulation the agents can perform a stopping maneuver or cross the street. Figure \ref{fig:crossing_stopping} illustrates such maneuvers with example images from the Daimler dataset \cite{Schneider_GCPR_2013}. For mapping the pedestrian detections to a vehicle-motion compensated ground plane, Schneider et al. used on-board sensors for velocity and yaw rate and a stereo camera system to compute the median disparity. Due to the non-linear observation model based on a perceptive camera model, an inevitable linearized extension for the Kalman and IMM filter observation models are required. Here, the observation uncertainty of the position sensor is assumed to be Gaussian distributed $r^{t} \sim \mathcal{N}(0, 0.01m)$ in the compensated reference system. Thus, the standard formulation of the Bayesian filters are well suited for this task. For the stopping maneuver or rather the event of deceleration till standing, a mean sojourn time of $1$ second with a standard deviation of 0.1 seconds is used. As long as a person moves in a straight line at a reasonably constant speed, their dynamics can be captured with a Kalman filter using a constant velocity model. During the maneuver, the relation to one fixed process model describing the dynamics fails due to an additional deceleration. Similar to Schneider et al. \cite{Schneider_GCPR_2013} or Kooij et al. \cite{Kooij_ECCV_2014}, the reference IMM filter is set up by combining two basic models, in particular, the constant velocity (CV) and the constant acceleration (CA) model. For avoiding side effects due to independent motions in different directions, see for example \cite{Becker_JMTA_2018}, only the crossing direction, from the vehicle perspective, the lateral motion is considered.
Following the aforementioned explanations, the IMM-RNN is compared to an IMM filter with two motion models (CV, CA), a Kalman filter with a single CV model, a Kalman filter with a single CA model, and as baseline to a linear interpolation. Also correspondingly to Schneider et al., the process noise $q$ is determined by $Q(t) = Q_{0}(t)q$, where $q \in \{\sigma_{CV}, \sigma_{CA} \}$ are spectral densities (continuous time variances) of the process noise, describing the changes in velocity or respectively in acceleration over a sampling period $\Delta t$ (CV: $\sqrt{Q_{22}}=\sqrt{\Delta t \cdot q} $; CA: $\sqrt{Q_{33}}=\sqrt{\Delta t \cdot q} $, see for example \cite{saerkkae_2013}). Based on this process noise model, the optimal process noise parameters for the different chosen filters (IMM filter (CV, CA), Kalman filter CV, CA) on the Daimler dataset are for the two IMM filter models $\sigma_{IMM, CV}=0.70,\sigma_{IMM, CA}=0.80$ and for the single Kalman filters $\sigma_{CV}=0.77$ and $\sigma_{CA}=0.44$ \cite{Schneider_GCPR_2013}. These parameters are consistent with the suggested practical setting in Bar-Shalom \cite{BarShalom_book_2002} and the chosen sojourn time for the simulation.\\
As mentioned above, a definition of maneuver classes for pedestrians is harder to establish than for vehicles. Hence, the main interest is here to detect the deviation from a standard behavior, and whether the pedestrian is in a \emph{normal} mode. A set of deviation in velocity, deceleration, along with the tangential ground truth trajectory is used to assign a maneuver label to a time step of a single trajectory. Thus, the RNN-IMM and IMM filter have a similar basic dynamic model set description. As the distribution over the trajectories for the RNN-IMM is captured with a Gaussian mixture model, the maneuver description for a single model can still be multi-modal. Since the IMM filter predicts a multi-modal distribution in form of a combination of the uni-modal model specific prediction, in the presented results the RNN-IMM is set to also only predict conditioned on a single maneuver class a uni-modal Gaussian distribution.\\

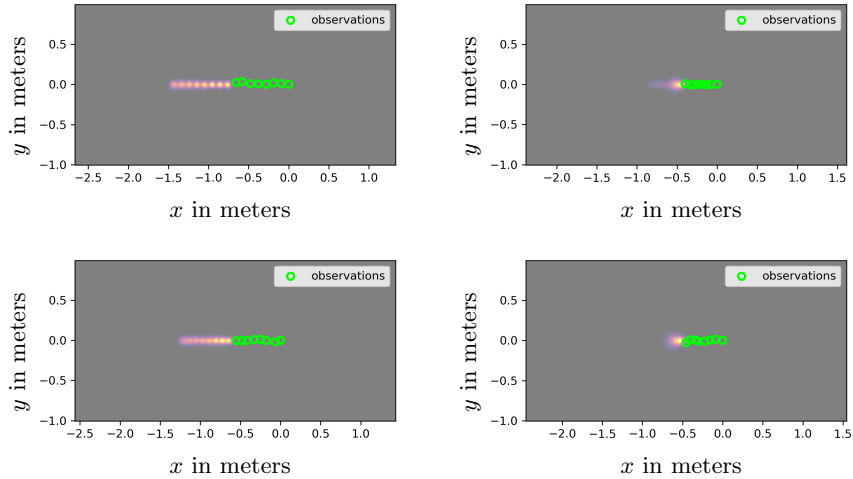
\begin{figure}[!h]
  \begin{center}
	\begin{tabular}{cc}				
				\begin{tikzpicture}		
					\node (image1) at (0,0) {	\includegraphics[width=.45\textwidth]{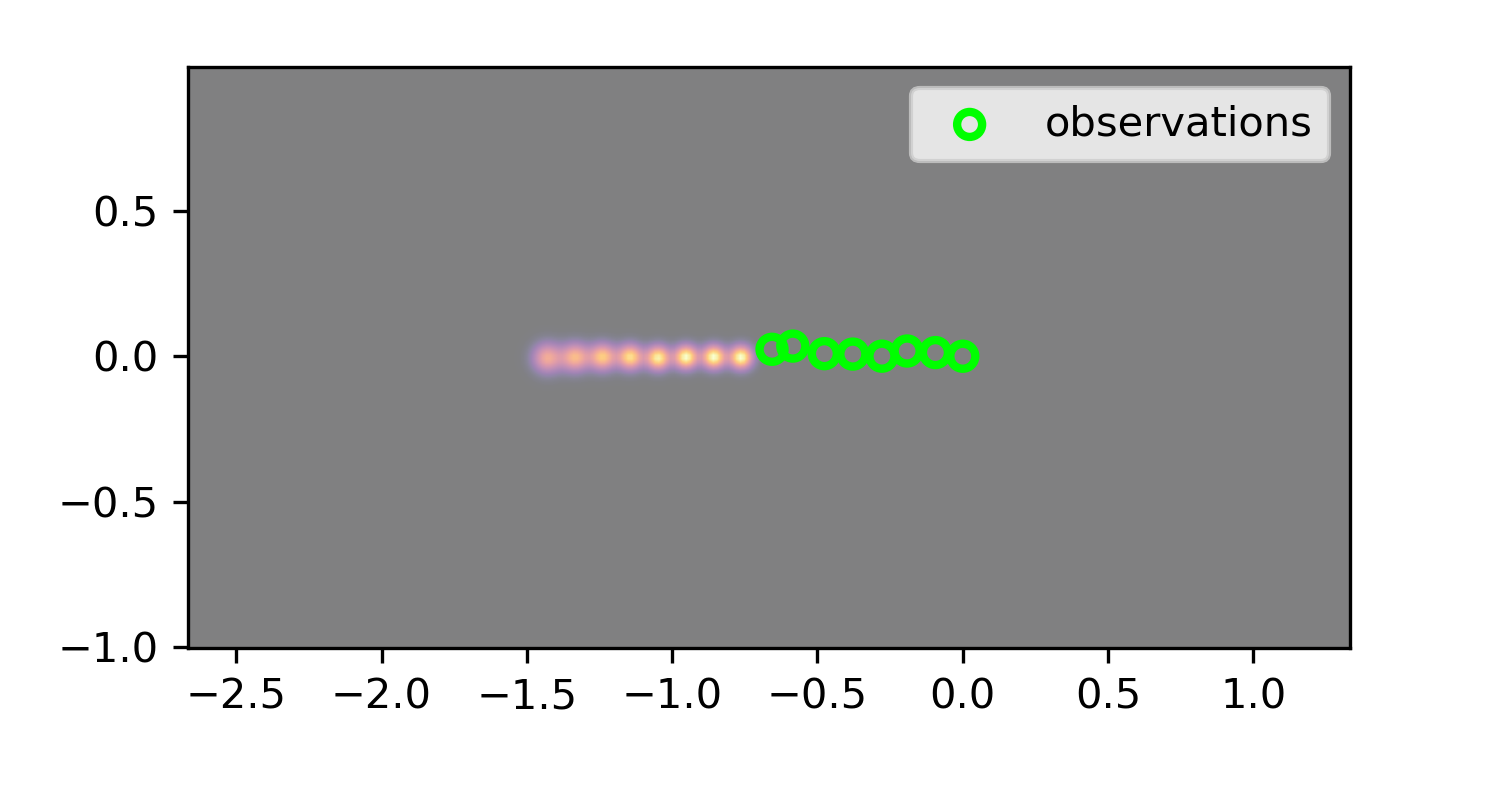}};			
					\node [below of = image1,node distance=1.5cm] (x){$x$ in meters};
					\node [left of = image1, node distance=2.8cm,rotate=90] (Y) {$y$ in meters};									
				\end{tikzpicture} &
				\begin{tikzpicture}
					\node (image2) at (0,0) {	\includegraphics[width=.45\textwidth]{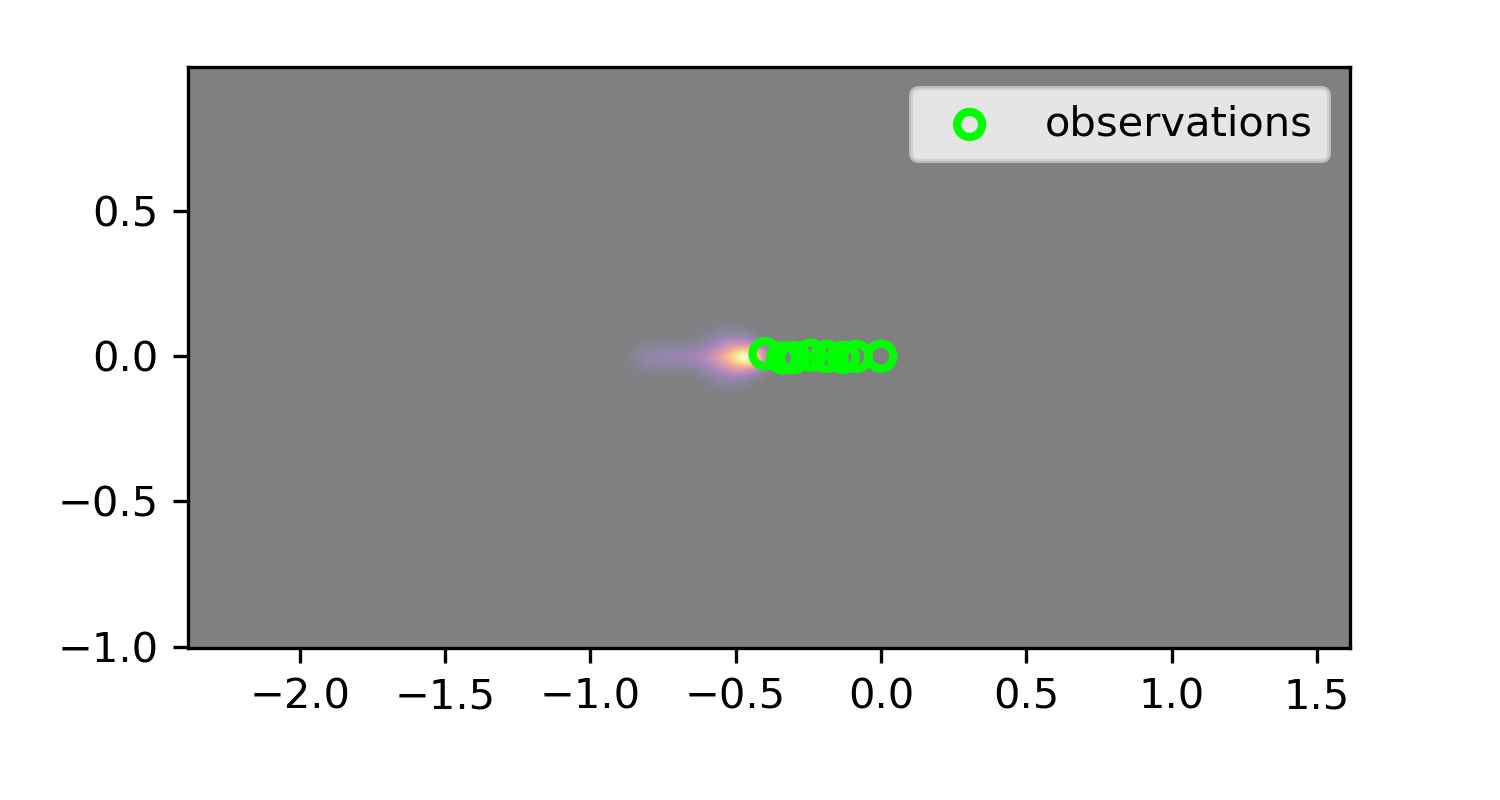}};
					\node [below of = image1,node distance=1.5cm] (x){$x$ in meters};
					\node [left of = image1, node distance=2.8cm,rotate=90] (Y) {$y$ in meters};			
				\end{tikzpicture}	
			   \\							
				\begin{tikzpicture}			
					\node (image3) at (0,0) {	\includegraphics[width=.45\textwidth]{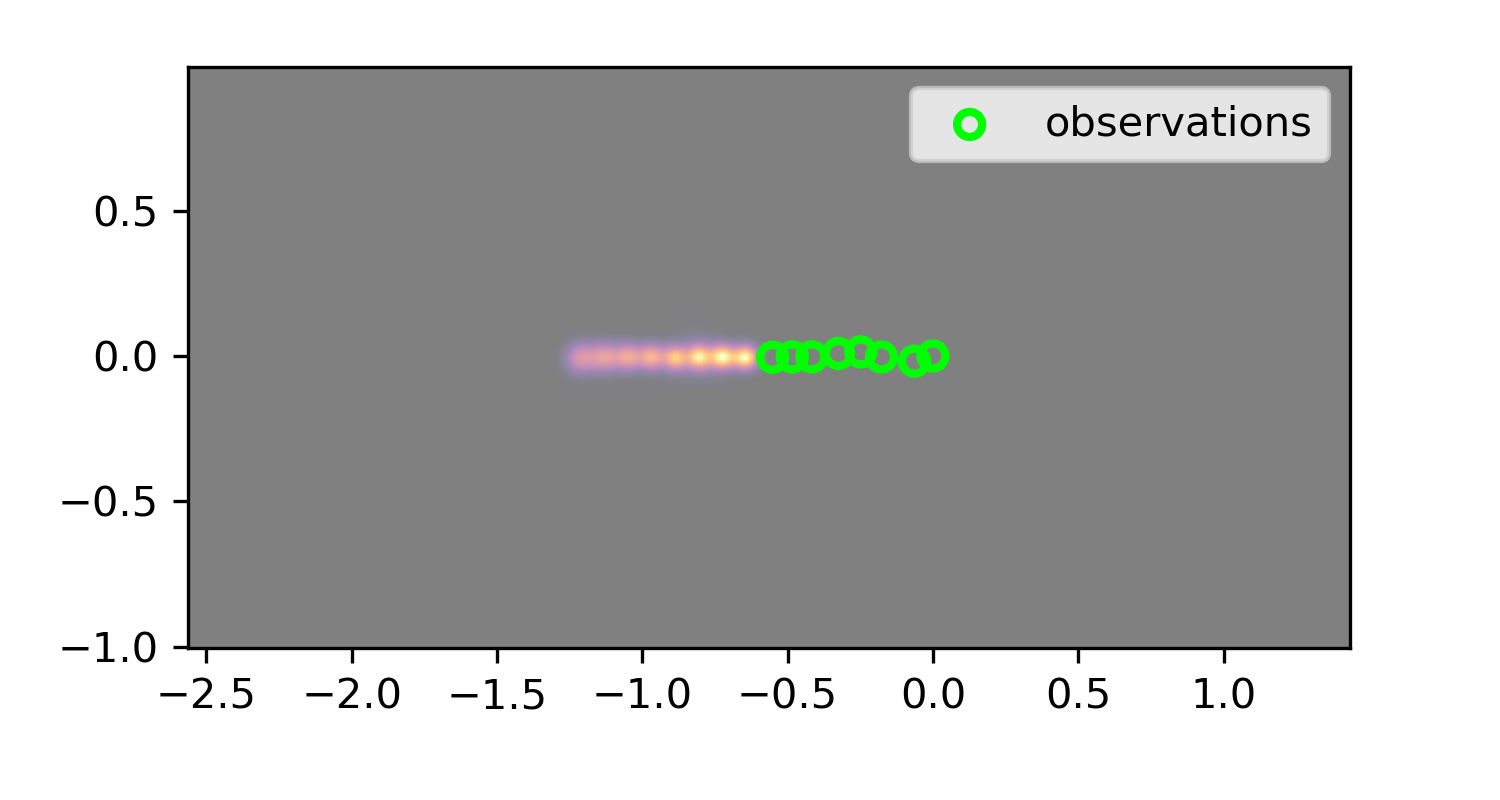}};			
					\node [below of = image1,node distance=1.5cm] (x){$x$ in meters};
					\node [left of = image1, node distance=2.8cm,rotate=90] (Y) {$y$ in meters};									
				\end{tikzpicture} &
				\begin{tikzpicture}
					\node (image4) at (0,0) {	\includegraphics[width=.45\textwidth]{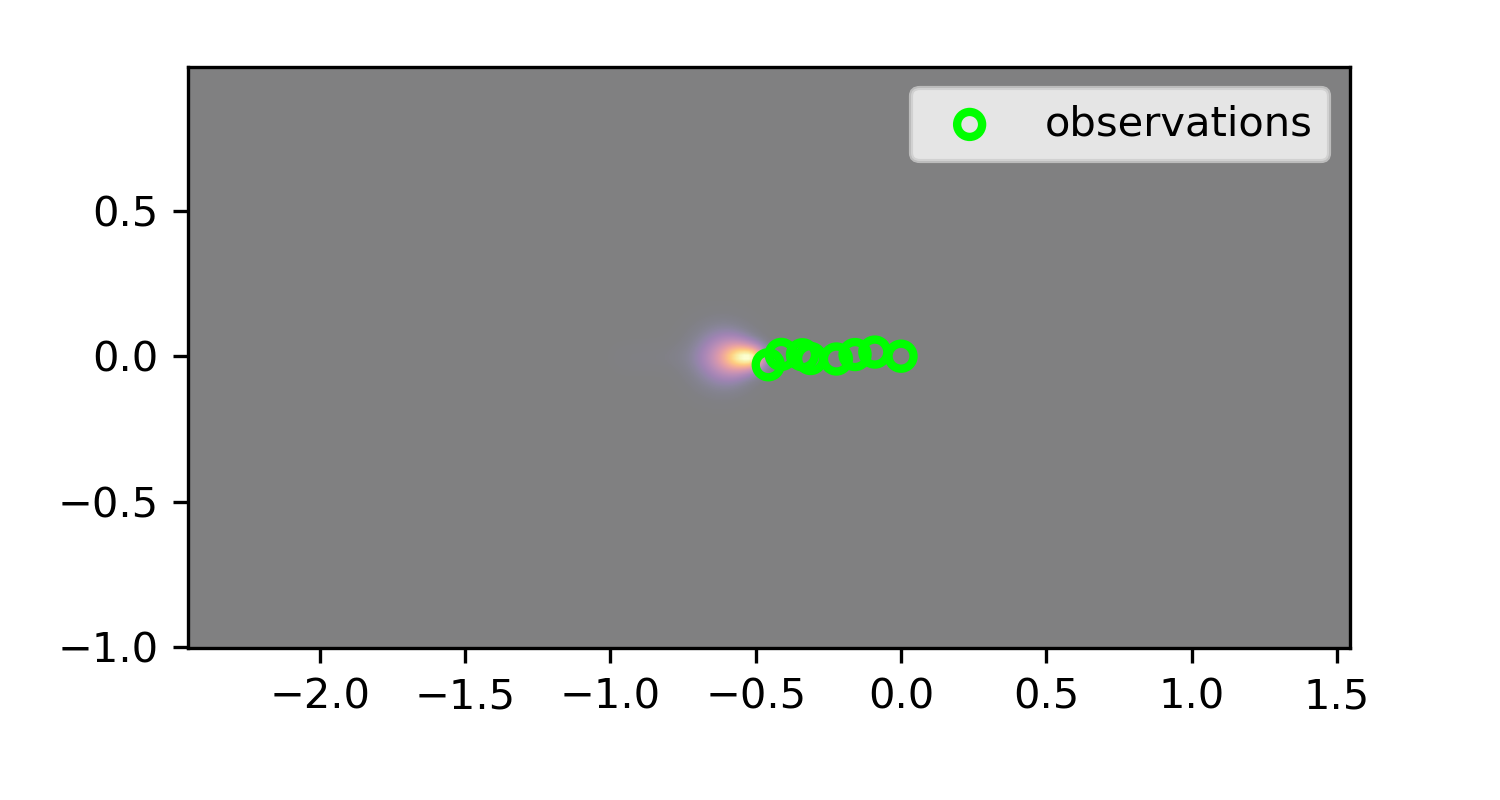}};
					\node [below of = image1,node distance=1.5cm] (x){$x$ in meters};
					\node [left of = image1, node distance=2.8cm,rotate=90] (Y) {$y$ in meters};			
				\end{tikzpicture}
  \end{tabular}						
	\end{center}
	\caption{\label{fig:rnn_man_results} Visualization of the predicted multi-modal distributions of future position as heatmap. (Left) Density plots for crossing or rather straight walking examples. (Right) Density plots for stopping examples in which the maximum of the predicted distribution is visible close to the last observation.} 
\end{figure}

\textbf{Implementation Details:} The model has been implemented using \emph{Tensorflow} \cite{tensorflow} and is trained for $2000$ epochs using ADAM optimizer \cite{Kingma_ICLR_2015} with a decreasing learning rate, starting from $0.01$ with a learning rate decay of $0.95$ and a delay factor of $1/10$. During the learning rate adaption, the number of epochs is multiplied by the delay factor. For the experiments, the RNN variant Long Short-Term Memory (LSTM) \cite{Hochreiter_NC_1997} is used.\\

\textbf{Results \& Analysis:} In figure \ref{fig:rnn_man_results}, predictions for two different preformed motion types are depicted for $8$ future positions weighted by the predicted maneuver probability. In the shown images the positions are normalized to start at the origin. The resulting multi-modal prediction is visualized as a heatmap. On the left, it can be seen that for a crossing sequence with straight walking the RNN-IMM mainly uses the corresponding straight walking model. On the right, where the deceleration started, the straight walking probability is visibly lower and the predicted distribution maximum is very close to the last observation.   
For the quantitative evaluation, $1000$ noisy trajectories have been synthetically generated, where $80$ percent are used for training and $20$ percent for the comparison to the recursive Bayesian filters. The results are summarized in table \ref{tab:RNN_IMM_results}.\\ 
The performance is compared with the final displacement error (FDE) (see for example \cite{Pellegrini_ICCV_2009}) of the lateral motion (from the vehicle perspective) for three different time horizons, in particular $8$ steps ($0.5s$), $12$ steps ($0.75s$) and $16$ steps ($1s$). These results show that the presented RNN-IMM is able to faster capture the change in dynamic for the synthetically generated data. In terms of the single motion models (CV vs. CA), one can observe the benefits for the CA in capturing the deceleration. The IMM filter combines both and shows an improvement. Hence, the aim of this paper is more on highlighting the relation between traditional multiple model approaches and the suggested RNN-based IMM filter surrogate, it should be mentioned that RNN-based approaches are designed to receive input data for every time step, whereas Bayesian filters are well suited for handling missing observations. Especially with such a short initialization time, this can be crucial. One argument towards a learning based RNN-IMM is that we only choose the maneuver definition based on deviation of standard straight walking. The engineering task of finding the best model set up for IMM filters and their extensions can lead to an improved behavior (see for example Keller et al. \cite{Keller_TITS_2014}) in specific maneuver situations, but is also very tedious to find a good setting. It should also be mentioned that recent work like the approaches of Kooij et al. \cite{Kooij_JCV_2018} show options how to further improve the prediction performance by including scene context and using more cues than pedestrian point kinematics (e.g. head orientation, gaze, body tilt, articulated body information).\\

\begin{table}[h!]
\caption{Results for the comparison between the proposed RNN-IMM and an IMM filter with two motion models (CV, CA), a Kalman filter with a single CV model, a Kalman filter with a single CA model, and using linear interpolation on the simulated maneuver situations. The prediction is done for $8$, $12$, and $16$ time steps conditioned on $8$ observations for a frame rate of $16$ fps.}
\label{tab:RNN_IMM_results}
\rowcolors{2}{blue!5}{gray!10}
\def\arraystretch{1.1}
\hspace*{-0cm}\begin{tabular}{ |c| c c | c c | c c|}
		\hline					
			\rowcolor{white!10} 
		\multicolumn{1}{|c|}{} & \multicolumn{2}{c|}{8/8} & \multicolumn{2}{c|}{8/12} & \multicolumn{2}{c|}{8/16 } \\
		\rowcolor{white!10} 
		\multicolumn{1}{|c|}{Approach} & \multicolumn{1}{c}{FDE [m]} & $\sigma_{\text{FDE}}$ [m] & \multicolumn{1}{c}{FDE [m]} & $\sigma_{\text{FDE}}$ [m]  & FDE [m]    &   \multicolumn{1}{c|}{$\sigma_{\text{FDE}}$ [m] }\\
				\hline
				\rowcolor{red!10}
				RNN-IMM  & 0.0309 & 0.0404 &	0.0427 & 0.0817 &  0.0517 & 0.0941\\
				\rowcolor{blue!10}				
				IMM filter (CV,CA) &  0.0674 & 0.0602  & 0.1188 & 0.1255  &  0.1862 & 0.1915\\
				\rowcolor{blue!10}	
				Kalman filter (CA) & 0.0796 & 0.0638 & 0.1575 &  0.1137  & 0.2386 & 0.1696\\
				\rowcolor{blue!10}
				Kalman filter (CV) & 0.1578 & 0.1601 & 0.2890 &  0.2965  & 0.4701 & 0.4700\\
				\rowcolor{blue!10}
				Linear interpolation & 0.1587 & 0.1610 &  0.2903 & 0.2978 & 0.4724 &  0.4718\\								
		   \hline
\end{tabular}\\
\centering
\end{table}

In summary, the presented RNN-IMM is able to also provide a confidence value $P(m_{i}|\mathcal{Z}) \overset{\wedge}{=} \alpha_{i}$ for the performed dynamic, but avoids modeling the dynamic transitions with a fixed transition probability matrix $P(m^{t}_{i}|m^{t-1}_{j})$. Similar to the provided mode probabilities of IMM filters, this can be used for further processing steps or rather applications (see for example \cite{Stierlin_ITSC_2012,Becker_SPIE_2015}). Further, instead of choosing the basic filter set, the prediction model is learned. In case there exists some well known model for describing the standard dynamic of the desired target, only deviations from the known dynamic can be used to define additional maneuver classes. This study on synthetically generated data shows, that by exploiting the connections between different views on maneuver prediction some perspectives on overcoming respective limitations can be gained.

\section{Conclusion}
\label{sec:conclusion}
In this paper, an RNN-encoder-decoder model, which can be interpreted as an IMM filter surrogate, has been presented. The RNN-IMM is able to jointly predict specific motion probabilities and corresponding distributions of future pedestrian trajectory. The model capabilities were shown on synthetic data that were reflecting typical pedestrian maneuvers. By conditioning on specific dynamic models or rather deviation from standard behavior, the model makes it possible to generate additional information in terms of an assigned maneuver probability similar to an IMM filter, but reduces the amount of explicit modeling of filter parameters (e.g. the dynamic transitions matrix). Thus, the presented RNN-IMM helps to reduce the amount of hard-coded engineering of traditional multiple model filter such as the IMM filter.  

%
%
%
\bibliographystyle{splncs04}
\bibliography{article}
\end{document}